\documentclass[10pt,twocolumn]{article}

\usepackage[letterpaper,margin=0.75in]{geometry}
\usepackage{caption}
\usepackage{cuted}
\usepackage{makecell} 
\usepackage[table]{xcolor} 
\usepackage{graphicx}
\usepackage{amsmath,amssymb}
\usepackage{booktabs}
\usepackage{multirow}
\usepackage{xcolor}
\usepackage{hyperref}

\title{Scaling 3D Generative Priors to Large-Scale Scene Meshes from Multi-View Images}

\author{
SangEun Lee, Wonseok Chae, Hoyoung Yoo, Geunyong Kim, NackWoo Kim, Hyeonjin Kim \\
Electronics and Telecommunications Research Institute
}

\date{}

\begin{document}

\maketitle

\begin{strip}
    \centering
    \includegraphics[width=\textwidth]{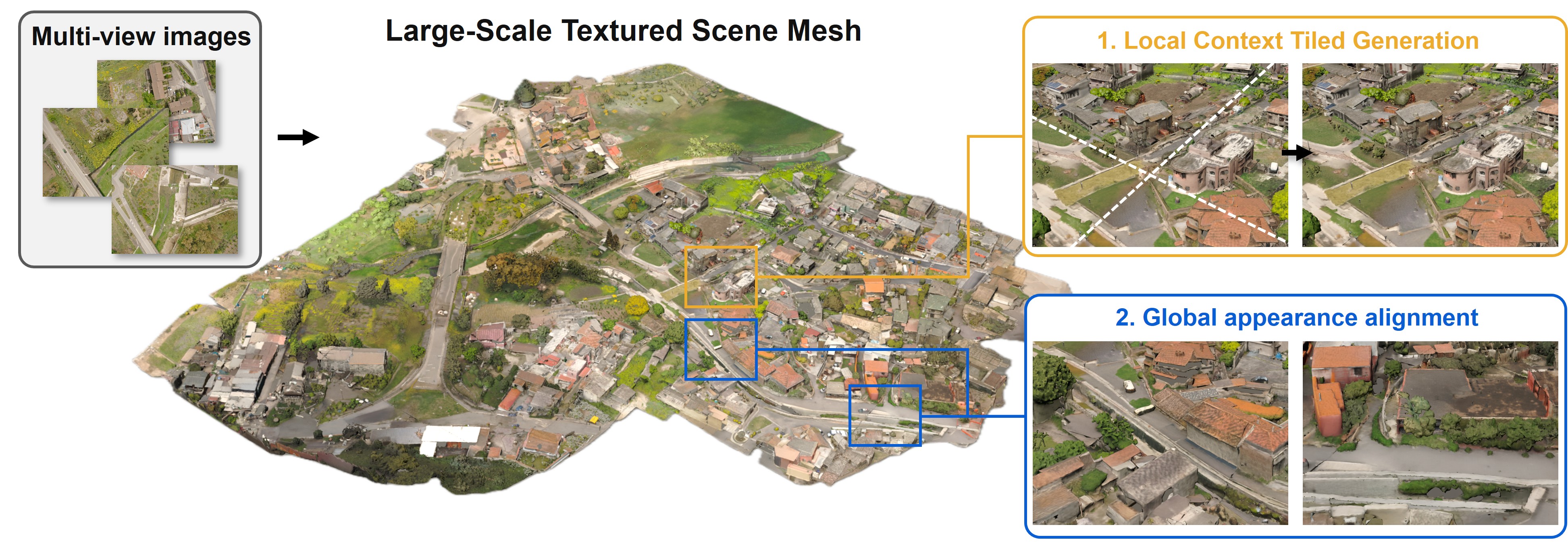}
    \captionof{figure}{Large-scale textured mesh generation from multi-view images with local continuity and global appearance consistency.}
    \label{fig:overview}
\end{strip}

\begin{abstract}
Pretrained 3D generative models produce detailed geometry and appearance but are primarily designed for object-centric generation within a limited spatial extent. Recent approaches address this limitation by partitioning large scenes into smaller spatial regions and applying pretrained 3D generative priors to each region. However, scaling tiled generation to large multi-view scenes makes it challenging to maintain local geometric continuity and global appearance consistency.

We present a training-free framework for large-scale textured mesh generation from multi-view images. Our key idea is to scale tiled generation to large scenes with increased spatial detail while coordinating generation both locally and globally. We introduce local context tiled generation to improve geometric continuity between neighboring regions and global appearance alignment to reduce appearance discrepancies across distant regions. An adaptive scene decomposition further determines the number of tiles according to the input scene geometry. Experiments demonstrate improved geometric and appearance fidelity over existing approaches while enabling fine-grained generation of large-scale scenes.
\end{abstract}

\section{Introduction}
\label{sec:intro} 

Recent 3D generative models have demonstrated impressive capabilities in
creating high-quality 3D meshes with detailed geometry and appearance from
a single image~\cite{xu2024instantmesh,yu2025fancy123}.
However, most of these models are trained primarily on individual objects and operate within a fixed spatial extent. Directly applying their fixed generative resolution to a much larger spatial extent inevitably reduces the level of geometric detail that can be represented.

Extend3D~\cite{yoon2026extend3d} addresses this limitation through tiled generation, extending object-level 3D generators to scene generation by dividing the scene into smaller spatial regions. Each region is generated at the native resolution of the pretrained model, preserving local detail while covering a larger scene. It coordinates neighboring regions by blending predictions in overlapping areas during denoising. This enables scene-level generation while retaining the pretrained model's native resolution.

However, scaling tiled generation to large environments with increased spatial detail requires observations over a broader spatial extent, which are difficult to capture with a single image. We therefore consider multi-view inputs that provide spatially distributed observations across the scene. Scaling tiled generation to this setting introduces additional challenges. 

First, as the number of tiles increases, jointly decoding the entire scene becomes increasingly memory-intensive. Tile-wise decoding alleviates this computational burden but can introduce geometric discontinuities at tile boundaries, making local geometric continuity difficult to maintain. Second, interactions limited to neighboring tiles cannot coordinate spatially distant regions, which can lead to inconsistent appearance across the scene.

To address these challenges, we present a training-free framework for large-scale textured mesh generation from multi-view images. Our \emph{Adaptive Tile Estimation} determines an appropriate scene decomposition based on the reconstructed scene geometry. We then introduce \emph{Local Context Tiled Generation}, which incorporates neighboring context during both latent denoising and tile-wise mesh decoding to improve geometric continuity across tile boundaries. Finally, \emph{Global Appearance Alignment} establishes global correspondences across distant regions to reduce scene-level appearance variations. Together, these components enable fine-grained tiled 3D generation of large-scale scenes from multi-view observations without additional training.

Our contributions are summarized as follows:
\begin{itemize}
    \item We present a training-free framework for large-scale textured mesh generation from multi-view images, with adaptive tile estimation that determines the scene decomposition according to the input scene geometry.
    \item We introduce a scalable local interaction scheme that incorporates neighboring information during both latent denoising and tile-wise mesh decoding, improving cross-tile geometric continuity without joint decoding of the entire scene.
    \item We propose global appearance alignment based on global scene correspondences, reducing appearance variations across spatially distant regions while preserving local details.
\end{itemize}

Extensive experiments demonstrate quantitative and qualitative improvements over existing approaches in geometric and appearance fidelity, while enabling fine-grained generation of large-scale scenes. Our code will be made publicly available.

\section{Related Work}
\subsection{3D Generative Models}
Recent advances in 3D generative modeling have substantially improved the fidelity and diversity of generated 3D assets~\cite{wang2023prolificdreamer,wu2024consistent3d,li2024grounded,
liu2023zero123,liu2023one2345,liu2024one2345pp,
hong2024lrm,wang2024crm,tochilkin2024triposr,
xu2024instantmesh,boss2024sf3d}. Pretrained 3D models such as TRELLIS~\cite{xiang2025structured}, Hunyuan3D~\cite{yang2024hunyuan3d,hunyuan3d2025hunyuan3d,zhao2025hunyuan3d,lai2025hunyuan3d}, and TripoSG~\cite{li2025triposg} employ diffusion or flow-based transformers over compact 3D latent representations~\cite{rombach2022high,lipman2023flow}, enabling high-quality geometry generation from image or text conditions. More recent approaches, including TRELLIS2~\cite{xiang2026native}, LATTICE~\cite{lai2026lattice}, and LATO~\cite{zhao2026lato}, further improve geometric detail and scalability through sparse structured representations, scalable latent spaces, and topology-aware mesh representations.

Despite these advances, most existing 3D generative models are primarily designed and trained for object-centric generation within bounded spatial domains. As a result, directly extending these models to spatially extended environments remains challenging. Extending these powerful object-level generative priors beyond their original spatial support therefore remains an important problem for large-scale 3D scene mesh generation.

\subsection{Large-Scale 3D Scene Generation}
Large-scale 3D scene reconstruction has been extensively studied with
neural representations such as Neural Radiance Fields (NeRF)~\cite{
mildenhall2020nerf,barron2022mipnerf360,tancik2022blocknerf,turki2022meganerf,rematas2022urbanradiance} and point-based representations such as 3D Gaussian Splatting (3DGS)~\cite{kerbl2023gaussiansplatting,lu2024scaffoldgs,lin2024vastgaussian,liu2024citygaussian}. More recently, several works have explored generative modeling at scene scale by extending generative models beyond individual objects. Existing approaches such as BlockFusion~\cite{wu2024blockfusion}, PDD~\cite{liu2024pyramid}, LT3SD~\cite{meng2025lt3sd}, and NuiScene~\cite{lee2025nuiscene} train scene-level generative models to synthesize spatially extended environments, often by decomposing scenes into local regions or adopting hierarchical representations. While these approaches demonstrate the potential of learned 3D scene generation, they typically require dedicated scene-scale training data and are often specialized to particular domains or scene categories.

Another line of work seeks to reuse pretrained object-level 3D generative priors for scene generation without extensive scene-specific training. SynCity~\cite{engstler2025syncity}, 3DTown~\cite{zheng2025constructing}, EvoScene~\cite{zheng2026self}, and Extend3D~\cite{yoon2026extend3d} generate larger environments by composing or spatially extending locally generated 3D content. In particular, Extend3D~\cite{yoon2026extend3d} enlarges the latent domain of a pretrained 3D generative model and performs denoising over overlapping local regions, enabling training-free generation beyond the spatial extent encountered during pretraining. However, coordination primarily through local overlapping regions can make it difficult to maintain geometric and appearance consistency as the number of tiles increases.

More recent approaches further explore scene generation under image observations. GenRecon~\cite{schmid2026genrecon} leverages a pretrained 3D generative prior together with posed multi-view images to reconstruct complete 3D scenes, while adapting the model to scene-level generation through additional training. In contrast, extending pretrained object-centric priors to large outdoor environments from calibrated multi-view observations without scene-specific training remains relatively underexplored.

\subsection{Tiled Diffusion for Large-Scale Generation}

Scaling pretrained diffusion models beyond their native resolution has been extensively studied in image generation~\cite{he2024scalecrafter,jimenez2023mixture}. MultiDiffusion~\cite{bartal2023multidiffusion} enables large-image synthesis by applying a pretrained diffusion model to overlapping local windows and aggregating their denoising predictions into a shared latent representation. Subsequent approaches such as SyncDiffusion~\cite{lee2023syncdiffusion} and DemoFusion~\cite{du2024demofusion} further improve coherence across local regions through synchronized denoising, global guidance, or multi-scale generation. These methods demonstrate that pretrained generative models can be extended to substantially larger domains without retraining.

A fundamental challenge of tiled generation is maintaining consistency among locally processed regions. Although overlapping-window aggregation can reduce visible discontinuities, local denoising operations have limited access to information outside their receptive regions, which can lead to inconsistent structures or appearances over long spatial ranges. This issue becomes particularly important for 3D scene generation, where geometric structures must remain continuous across neighboring regions while appearance should remain consistent over distant regions. These observations motivate both local information exchange and global coordination when extending pretrained 3D generative models to large spatial domains.
 
\section{Preliminaries}
\label{sec:preliminaries} 
\subsection{Pretrained 3D Generative Priors} 

Our framework builds upon TRELLIS2~\cite{xiang2026native} as a pretrained 3D generative
prior. TRELLIS2 generates a 3D asset through a sequence of sparse structure,
geometry, and material generation.

Given an input image $I$, TRELLIS2 first performs \emph{sparse structure generation}, which predicts the active voxels that define the spatial support for subsequent generation. Conditioned on the predicted structure, \emph{geometry generation} produces geometry latents at the active voxels, followed by \emph{material generation}, which synthesizes material latents aligned with the generated geometry. These stages employ diffusion transformers (DiTs) trained with a flow matching paradigm and conditioned on the input image.

The generation process can be summarized as
\begin{equation}
    S = \mathcal{G}_{\mathrm{ss}}(I), \qquad
    Z_g = \mathcal{G}_{\mathrm{geo}}(S,I), \qquad
    Z_m = \mathcal{G}_{\mathrm{mat}}(S,Z_g,I)
\end{equation}
where $S$ denotes the sparse structure, and $Z_g$ and $Z_m$ denote the
geometry and material latents, respectively.

The generated geometry and material latents are decoded by their respective SC-VAE decoders into O-Voxel, a sparse voxel representation that contains local geometry and material attributes. The decoded geometry provides the Flexible Dual Grid features used to reconstruct the mesh surface. The decoded material attributes are then assigned to the reconstructed surface.

TRELLIS2 is pretrained primarily on object-level 3D data. Its finite generative resolution limits the spatial extent that can be represented while preserving fine geometric details. This limitation motivates spatially decomposed generation for large-scale scenes.

\begin{figure*}
    \centering
    \includegraphics[width=\linewidth]{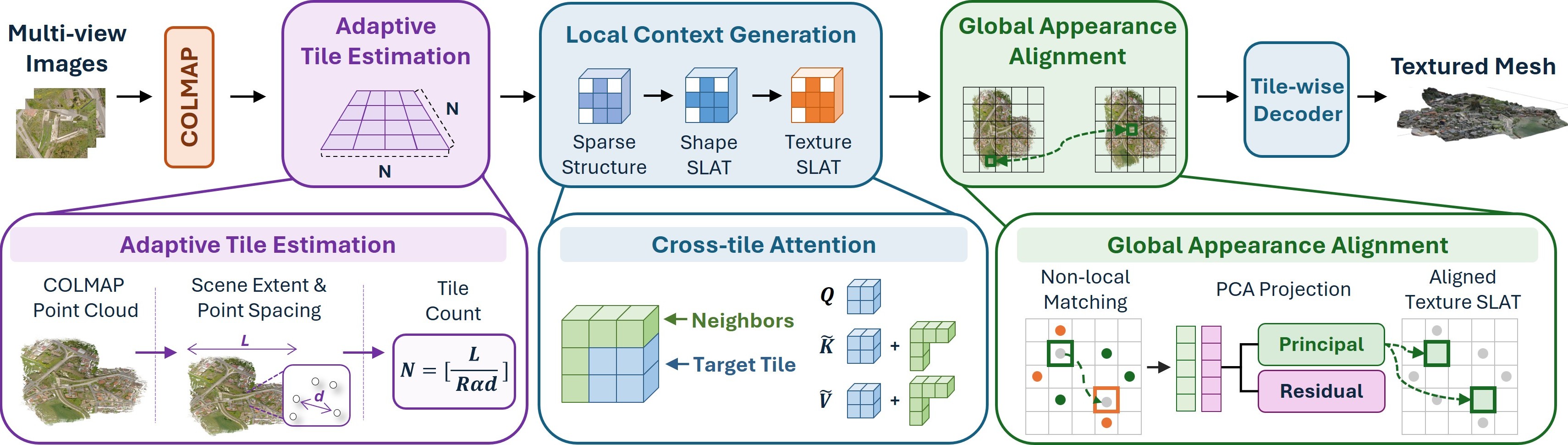}
    \caption{Overview of our framework. Our method combines geometry-adaptive scene decomposition, local context tiled generation, and global appearance alignment to generate coherent large-scale 3D scene meshes from multi-view images. The tile-wise decoding process shown in the upper right is detailed in Fig.~\ref{fig:method_decoding}.}
    \label{fig:method}
\end{figure*} 

\subsection{Tiled 3D Scene Generation}
\label{sec:prelim_extend3d}

Our work starts from the tiled generation formulation introduced by
Extend3D~\cite{yoon2026extend3d}, a training-free framework for large-scale 3D scene
generation from a single image. Given an input image, Extend3D generates a larger 3D scene by dividing it into overlapping local regions, each processed using a pretrained object-centric 3D generator. Let $\Omega$ denote the target scene domain with spatial extent $W \times L$. It divides the scene into $n$ and $m$ regions along the width and length dimensions, respectively. The tiled representation is defined as
\begin{equation}
\Omega =
\bigcup_{i=1}^{n}
\bigcup_{j=1}^{m}
\Omega_{i,j},
\end{equation}
where $\Omega_{i,j}$ denotes the local region at grid position $(i,j)$, with adjacent regions spatially overlapping to facilitate cross-tile coordination.
Each tile is generated at the native resolution of the pretrained model, allowing larger scenes to be generated while retaining the native generative resolution within each local region. Predictions from neighboring tiles are blended in their overlapping regions to reduce boundary discrepancies.

We adopt this tiled formulation as our starting point and extend it to
\emph{large-scale scene generation from multi-view observations}. Scaling this formulation requires addressing several issues. First, the appropriate decomposition scale varies with the scene extent and sampling density, making a fixed tiling strategy suboptimal. Second, as the number of tiles increases, jointly decoding the entire latent representation becomes increasingly memory-intensive. Tile-wise decoding alleviates this bottleneck but removes cross-tile interactions during decoding, potentially introducing discontinuities at tile boundaries. Third, the blending of overlapping regions only connects neighboring tiles and does not enforce appearance consistency between distant parts of the scene.

\section{Method}
\label{sec:method}
\subsection{Overview}
Given a set of multi-view images $\{I_k\}_{k=1}^{K}$, our goal is to
generate a detailed textured mesh of a large-scale scene by leveraging a pretrained 3D generative prior without additional training. As illustrated in Fig.~\ref{fig:method}, we first recover camera poses and a 3D point cloud from the input images using COLMAP~\cite{schonberger2016structure,schonberger2016pixelwise}. The reconstructed point cloud provides the basis for scene decomposition, while the camera poses are used to associate the input observations with their corresponding scene regions.

Our framework consists of three main components.
First, \textbf{Adaptive Tile Estimation} automatically determines the
number of tiles based on the spatial extent and sampling density of the
reconstructed scene (Sec.~\ref{sec:adaptive_decomposition}).
Given the resulting tiles, \textbf{Local Context Tiled Generation}
enables scalable tile-wise generation while incorporating context from
adjacent regions to reduce discontinuities between neighboring tiles
(Sec.~\ref{sec:neighbor_generation}).
Finally, \textbf{Global Appearance Alignment} establishes correspondences
between spatially distant regions and selectively aligns their texture
latent features to improve appearance consistency across the scene
(Sec.~\ref{sec:appearance_alignment}).
The resulting tile representations are assembled into the final
textured scene mesh.

\subsection{Adaptive Tile Estimation}
\label{sec:adaptive_decomposition} 

A fixed scene decomposition does not account for variations in scene size and sampling density. Using too few tiles increases the spatial region represented by each voxel, potentially reducing geometric detail. We therefore determine the number of tiles according to the spatial extent and sampling density of the reconstructed scene.

Let $\mathcal{P}$ denote the dense point cloud reconstructed from the input multi-view images using COLMAP. We define the horizontal scene extent along the two spatial axes as
\begin{equation}
W = \operatorname{extent}_x(\mathcal{P}),
\qquad
L = \operatorname{extent}_y(\mathcal{P}).
\end{equation}
To characterize the sampling density of the observed geometry, we compute the median $k$-nearest-neighbor distance over $\mathcal{P}$:
\begin{equation}
d =
\operatorname{median}_{\mathbf{p}\in\mathcal{P}}
d_k(\mathbf{p}),
\end{equation}
where $d_k(\mathbf{p})$ denotes the distance from point $\mathbf{p}$ to its $k$-th nearest neighbor. We use $k=5$ in all experiments. The point spacing $d$ serves as a proxy for the spatial detail supported by the reconstructed geometry.

Suppose the scene is decomposed into an $n\times m$ grid, where $n$ and $m$ denote the number of tiles along the width and length dimensions, respectively. Since each tile contains $R$ voxels along each spatial axis, the effective voxel sizes are
\begin{equation}
v_x = \frac{W}{Rn},
\qquad
v_y = \frac{L}{Rm}.
\end{equation}
We constrain these voxel sizes according to the observed point spacing:
\begin{equation}
v_x \leq \alpha d,
\qquad
v_y \leq \alpha d,
\end{equation}
where $\alpha$ controls the allowable voxel size relative to the observed geometry. This yields
\begin{equation}
n =
\left\lceil \frac{W}{R\alpha d} \right\rceil,
\qquad
m =
\left\lceil \frac{L}{R\alpha d} \right\rceil.
\end{equation}

Following the 1024³ generation setting of TRELLIS2, we set the sparse latent resolution to R=64. We empirically set $\alpha=5$ on a reference scene and keep it fixed for all experiments. This formulation allocates more tiles to larger or more densely sampled scenes while avoiding unnecessarily fine decomposition for sparsely sampled geometry.

\subsection{Local Context Tiled Generation}
\label{sec:neighbor_generation}
Existing tiled generation blends predictions in overlapping regions during
denoising and jointly decodes the resulting scene representation.
While joint decoding provides context across tile boundaries, its memory
cost increases with the number of tiles, limiting its scalability to finer
scene decompositions.
We instead process each tile independently during decoding, reducing the peak memory requirement and enabling generation with a larger
number of tiles.

However, tile-wise decoding removes the cross-tile context available in
joint decoding and can introduce geometric discontinuities at tile
boundaries. Moreover, overlap-based blending during denoising only combines
predictions in shared regions and does not directly provide neighboring
features to the internal processing of each tile. We therefore introduce local cross-tile context into both latent generation and mesh decoding.

\paragraph{Cross-Tile Attention.}
\label{para:cross_tile_attention}
We incorporate neighboring context directly into the self-attention layers
of the pretrained denoiser. For each tile $i$, we collect key and value
features from the adjacent boundary regions of its neighboring tiles.
The queries remain local to the current tile, while the keys and values are
augmented as
\begin{equation}
    \widetilde{K}_i = [K_i; K_i^{\mathrm{nbr}}],
    \qquad
    \widetilde{V}_i = [V_i; V_i^{\mathrm{nbr}}],
\end{equation}
where $K_i^{\mathrm{nbr}}$ and $V_i^{\mathrm{nbr}}$ denote the key and value
features collected from neighboring boundary regions. The attention is then
computed using $Q_i$, $\widetilde{K}_i$, and $\widetilde{V}_i$.

Unlike overlap-based prediction blending, which combines predictions from
neighboring tiles, our approach introduces neighboring context directly
within the denoiser. Restricting the additional features to adjacent
boundaries provides cross-tile context without jointly processing the
complete scene representation.

\begin{figure}
    \centering
    \includegraphics[width=\linewidth]{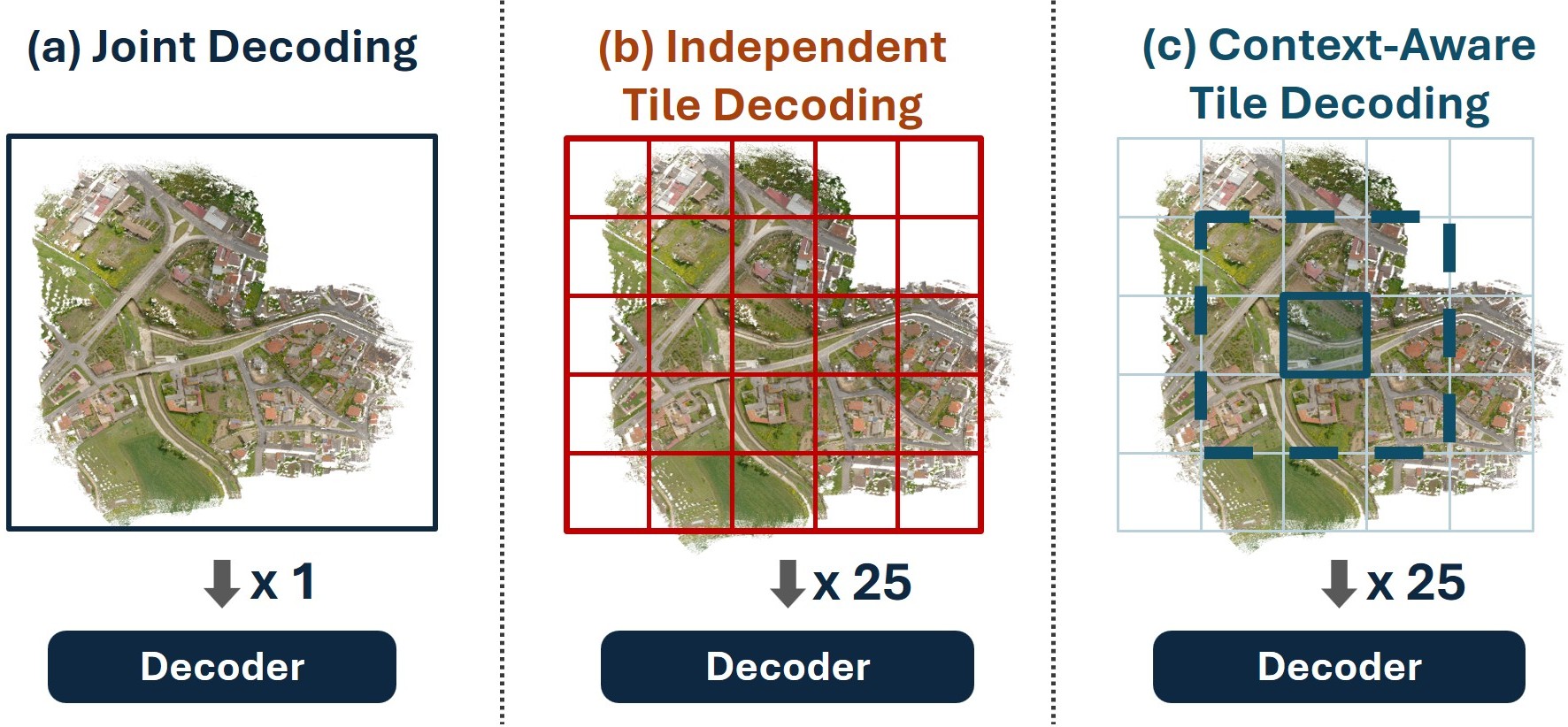}
    \caption{Comparison of decoding strategies. Our context-aware tile decoding incorporates neighboring halo features during tile-wise decoding, providing cross-tile context without jointly decoding the entire scene.}
    \label{fig:method_decoding}
\end{figure}

\paragraph{Context-Aware Tile Decoding.}
\label{para:context_tile_decoding}
Cross-tile context is also important during mesh reconstruction.
After generation, the sparse latent representations of individual tiles
are aligned in a common scene coordinate system.
Independently decoding each tile prevents the decoder from accessing latent
features across tile boundaries, which can introduce discontinuities in the
reconstructed geometry. As shown in Fig~\ref{fig:method_decoding}, we therefore augment each tile with a narrow halo of latent voxels from its
spatial neighbors before decoding.
The augmented tile is passed through the pretrained decoder, while only the
output corresponding to its core region is retained.
This provides neighboring context during decoding while maintaining the
memory efficiency of tile-wise processing.

\subsection{Global Appearance Alignment}
\label{sec:appearance_alignment}

Local cross-tile interactions provide context between adjacent tiles but
do not directly constrain the appearance of distant regions. As a result,
corresponding surfaces observed from different local views may exhibit inconsistent appearance. We address this by first identifying
global correspondences across the scene and then selectively aligning
their texture-SLAT features.

\paragraph{Global Voxel Correspondence.}
Following prior work that lifts multi-view image features into 3D for correspondence estimation~\cite{zhang2023tale,zhu2025densematcher}, we associate each active texture-SLAT voxel with image features aggregated from its visible views. 
For each active texture-SLAT voxel $p$ at position $\mathbf{x}_p$, we
project it onto the visible input views and average the corresponding DINO
features:
\begin{equation}
    f_p =
    \frac{1}{|\mathcal{V}(p)|}
    \sum_{k \in \mathcal{V}(p)}
    F_k\bigl(\pi_k(\mathbf{x}_p)\bigr),
\end{equation}
where $F_k$ denotes the DINO feature map extracted from image $I_k$,
$\mathcal{V}(p)$ denotes the set of input views in which $p$ is visible,
and $\pi_k$ projects the 3D position $\mathbf{x}_p$ onto view $k$.

We use $f_p$ as a view-aggregated descriptor and retrieve the $K$ most
similar voxels from other tiles using cosine similarity:
\begin{equation}
    \mathcal{N}(p)
    =
    \operatorname{TopK}_{q:T(q)\neq T(p)}
    \operatorname{cos}(f_p,f_q),
\end{equation}
where $T(p)$ denotes the tile containing voxel $p$.

\definecolor{best}{RGB}{210,232,255}
\definecolor{second}{RGB}{235,244,252}

\begin{table*}[t]
\centering
\caption{Quantitative comparison with existing 3D generation and reconstruction methods.
Best results are shown in \textbf{bold}, and second-best results are \underline{underlined}.}
\label{tab:main_results}

\small
\setlength{\tabcolsep}{8pt}
\begin{tabular}{lc|cccc|cccc}
\toprule
\multirow{2}{*}{Method}
& \multirow{2}{*}{\# Views}
& \multicolumn{4}{c|}{Synthetic}
& \multicolumn{4}{c}{Urban} \\
\cmidrule(lr){3-6}
\cmidrule(lr){7-10}

&
& LPIPS $\downarrow$
& SSIM $\uparrow$
& CD $\downarrow$
& F-score $\uparrow$
& LPIPS $\downarrow$
& SSIM $\uparrow$
& CD $\downarrow$
& F-score $\uparrow$ \\
\midrule
Hunyuan3D-2.1~\cite{hunyuan3d2025hunyuan3d}
& Single
& 0.648
& 0.308
& \cellcolor{second}\underline{0.016}
& \cellcolor{best}\textbf{0.934}
& 0.728
& 0.162
& 0.066
& 0.663 \\

TRELLIS2~\cite{xiang2026native}
& Single
& 0.482
& 0.443
& 0.023
& 0.873
& \cellcolor{second}\underline{0.676}
& 0.206
& 0.008
& 0.959 \\

EvoScene~\cite{zheng2026self}
& Single
& 0.491
& 0.486
& 0.023
& 0.834
& 0.722
& 0.174
& 0.011
& 0.920 \\

Extend3D~\cite{yoon2026extend3d}
& Single
& 0.412
& 0.560
& 0.024
& 0.825
& 0.778
& \cellcolor{second}\underline{0.250}
& \cellcolor{second}\underline{0.006}
& \cellcolor{second}\underline{0.976} \\

\midrule

GenRecon~\cite{schmid2026genrecon}
& Multi
& \cellcolor{second}\underline{0.356}
& \cellcolor{second}\underline{0.635}
& 0.037
& 0.652
& 0.747
& 0.213
& 0.041
& 0.695 \\

\textbf{Ours}
& Multi
& \cellcolor{best}\textbf{0.241}
& \cellcolor{best}\textbf{0.723}
& \cellcolor{best}\textbf{0.015}
& \cellcolor{second}\underline{0.925}
& \cellcolor{best}\textbf{0.634}
& \cellcolor{best}\textbf{0.412}
& \cellcolor{best}\textbf{0.002}
& \cellcolor{best}\textbf{0.999} \\

\bottomrule
\end{tabular}
\end{table*}

\paragraph{Selective Latent Alignment.}
Directly modifying the full texture-SLAT feature may unnecessarily alter
local appearance information. We therefore align only the principal feature components that capture dominant appearance variations, while preserving the residual components containing local details.
Given a texture feature $z_p$, its projection onto the principal subspace is
\begin{equation}
    z_p^{P}
    =
    \mu + U_dU_d^\top(z_p-\mu),
\end{equation}
where $\mu$ denotes the mean of the scene-level texture-SLAT features and
$U_d$ contains the top $d$ principal directions obtained by PCA.

For each voxel $p$, we compute a similarity-weighted consensus from its
matched voxels. We define the weight for a matched voxel
$q \in \mathcal{N}(p)$ as
\begin{equation}
    w_{pq}
    =
    \max\left(0,\operatorname{cos}(f_p,f_q)\right).
\end{equation}
The consensus feature in the principal subspace is
\begin{equation}
    \bar{z}_p^{P}
    =
    \frac{
        \sum_{q\in\mathcal{N}(p)} w_{pq} z_q^{P}
    }{
        \sum_{q\in\mathcal{N}(p)} w_{pq}
    }.
\end{equation} 
Voxels with no positive-weight correspondences are left unchanged.

We update the original texture feature using this consensus:
\begin{equation}
    z'_p
    =
    z_p + \alpha
    \left(\bar{z}_p^{P}-z_p^{P}\right),
\end{equation}
where $\alpha$ controls the alignment strength. This update modifies only the principal-subspace component while preserving the residual $z_p-z_p^{P}$.


\begin{figure*}
    \centering
    \includegraphics[width=\linewidth]{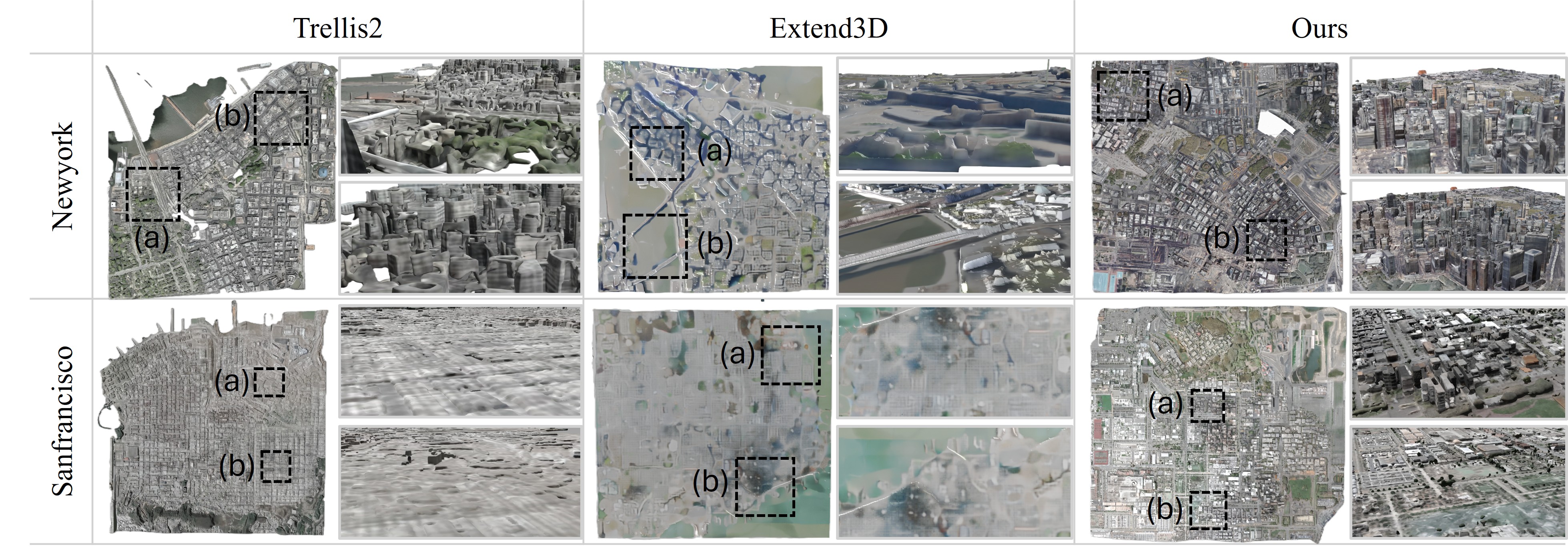}
    \caption{Qualitative comparison on the NewYork and SanFrancisco scenes. (a) and (b) show zoomed-in views of the upper-right and lower-right regions of each cell, respectively.}
    \label{fig:main_results}
\end{figure*}

\section{Experiments}  

\subsection{Main Results} 
We evaluate our method on UrbanScene3D~\cite{lin2022capturing}, a large-scale
dataset for urban scene reconstruction.
We use seven scenes covering diverse spatial extents and urban structures:
all four synthetic benchmark scenes (\emph{School}, \emph{Town},
\emph{Bridge}, and \emph{Castle}) and three large-scale urban scenes
(\emph{New York}, \emph{San Francisco}, and \emph{Chicago}).

We evaluate both the appearance and geometry of the generated textured
meshes.
For appearance evaluation, we render the generated meshes from the
corresponding camera viewpoints and measure LPIPS~\cite{zhang2018unreasonable} and SSIM~\cite{wang2004image}.
For geometry evaluation, we compare the generated meshes with the ground-truth geometry provided by the dataset using Chamfer Distance and F-score with a distance threshold of 0.05.

\paragraph{Quantitative comparison.} 
Table~\ref{tab:main_results} reports the quantitative results on the
synthetic and urban scenes of UrbanScene3D.
Our method demonstrates superior overall performance compared with existing approaches.
In terms of appearance, our method obtains the lowest LPIPS and the highest
SSIM, indicating that the generated scenes better reproduce the
input observations than the compared methods. Our method also demonstrates strong performance in Chamfer Distance and F-score, indicating improved geometric agreement with the ground-truth scene geometry.

The single-image generation methods, including TRELLIS2, Hunyuan3D-2.1,
EvoScene, and Extend3D, show lower performance when reconstructing these
large-scale scenes from limited observations.
In contrast, our method leverages spatially distributed multi-view
observations together with scalable tiled generation, resulting in more
accurate appearance and geometry across the scene.
GenRecon, despite utilizing multi-view observations, also performs worse
than our method on the outdoor synthetic scenes.
As GenRecon is primarily developed for indoor scene reconstruction, this
performance gap may partly reflect the domain difference between its target
setting and the large-scale outdoor scenes considered in our evaluation. In contrast, our training-free framework effectively transfers the pretrained 3D generative prior to large-scale outdoor scenes without scene-specific adaptation.

\definecolor{ours}{RGB}{235,244,252} 
\subsection{Qualitative Comparison}

Figure~\ref{fig:main_results} presents qualitative comparisons on the New York and San Francisco scenes against TRELLIS2 and Extend3D.
Since both baselines take a single image as input, we provide a top-view image of each scene as their input.
We visualize both the top view of the generated scene and enlarged regions to examine local geometric and appearance details.

Our method better preserves the overall scene structure while producing more detailed buildings.
In particular, the enlarged regions show greater variation in building heights and geometry that more closely reflects the input observations.
Our results also retain finer facade and texture details, whereas the baselines tend to produce smoother geometry or less distinctive appearances.
These results demonstrate that leveraging multi-view observations together with our tiled generation framework enables more faithful and detailed reconstruction of large-scale urban scenes.

\begin{figure*}
    \centering
    \includegraphics[width=0.95\linewidth]{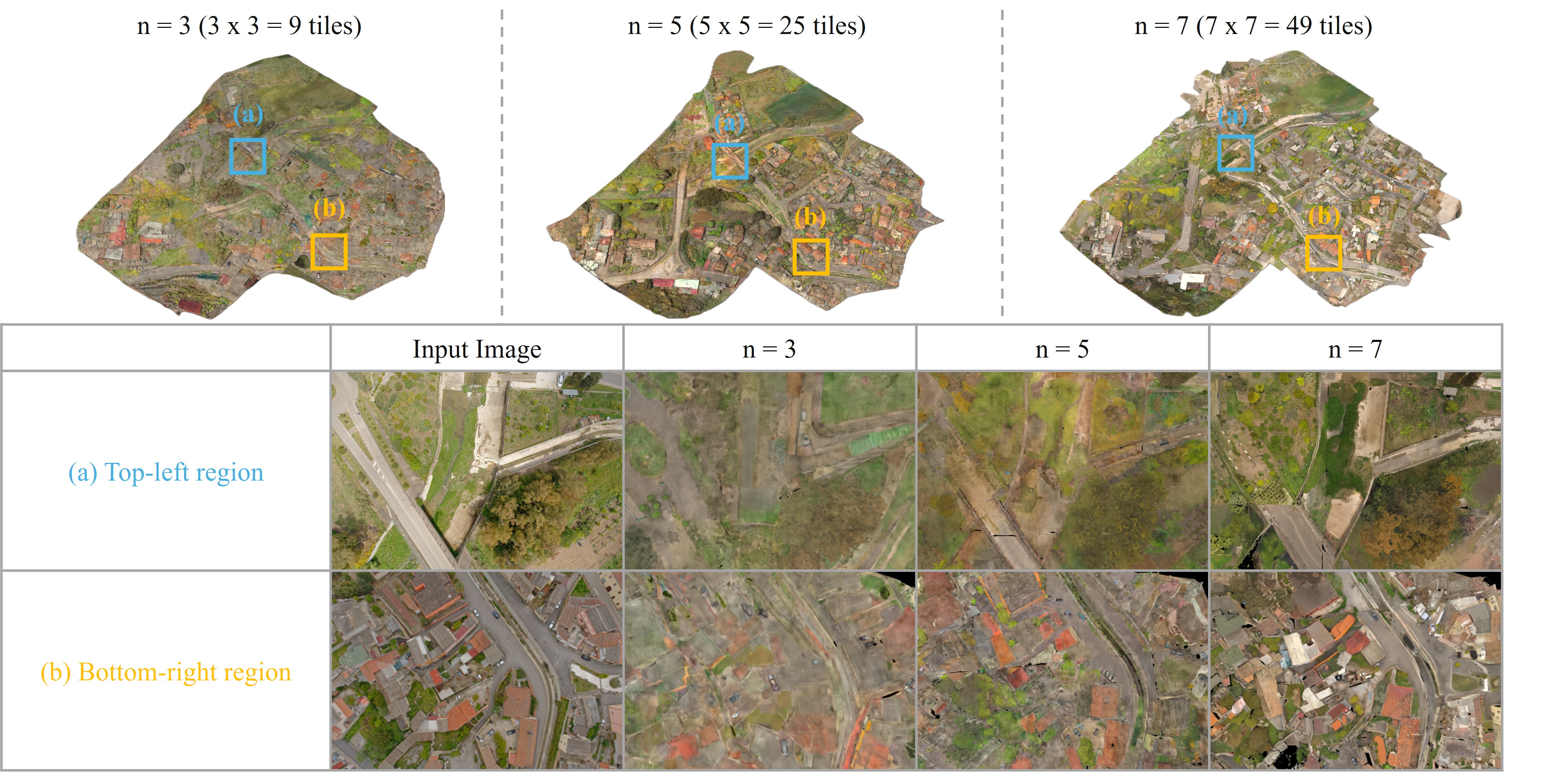}
    \caption{Large-scale scene generation with increasing tile resolution
($n=3,5,7$) over a fixed scene extent. Finer decompositions allocate the
pretrained generator's native resolution to smaller spatial regions,
revealing progressively finer geometric and appearance details.}
    \label{fig:scaling}
\end{figure*}

\begin{figure*}
    \centering
    \includegraphics[width=\linewidth]{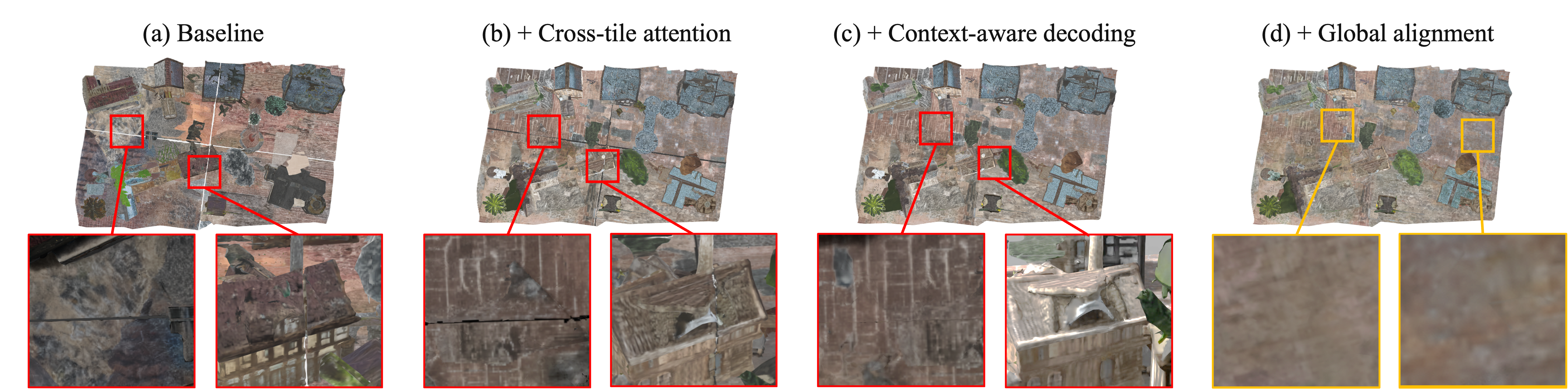}
    \caption{Qualitative ablation of our method. We progressively add (b) cross-tile attention, (c) context-aware decoding, and (d) global alignment to (a) the baseline, improving local continuity and global appearance consistency.
}
    \label{fig:ablation}
\end{figure*}  

\subsection{Scalability Analysis}

We evaluate the scalability of our method by increasing the tile resolution
from $n=3$ to $n=5$ and $n=7$ while maintaining the same spatial extent of
the Town scene. As shown in Fig.~\ref{fig:scaling}, increasing the number of
tiles allocates the native resolution of the pretrained 3D prior to
progressively smaller spatial regions, allowing finer geometric structures
and appearance details to be represented.

Table~\ref{tab:scalability} quantitatively shows the effect of increasing
the tile resolution. As $n$ increases from 3 to 7, the number of generated
SLAT voxels increases from 74,990 to 410,868, while the resulting mesh
complexity increases from 1.13M to 5.67M faces. At the same time, LPIPS
decreases from 0.683 to 0.605, indicating improved agreement between the
rendered scene and the reference images. These results show that increasing
the number of tiles provides a higher-resolution scene representation of the same scene area.

\begin{table}[t]
\centering
\caption{Scalability analysis with increasing numbers of tiles on the Town scene. 
The scene extent is fixed across all settings.}
\label{tab:scalability}
\small
\setlength{\tabcolsep}{6pt}

\begin{tabular}{c|c|ccc}
\toprule
$n$ & \# Tiles
& LPIPS $\downarrow$
& \# Faces
& \# SLAT Voxels \\
\midrule

3 & 9  & 0.683 & 1,132,659 & 74,990 \\

5 & 25 & 0.620 & 3,038,380 & 195,223 \\

\rowcolor{ours}
\textbf{7} & \textbf{49} & \textbf{0.605} 
& \textbf{5,677,351} & \textbf{410,868} \\

\bottomrule
\end{tabular}
\end{table}

\subsection{Ablation Study}
\begin{table}[t]
\centering
\caption{Ablation study of the proposed components.}
\label{tab:ablation}
\small
\setlength{\tabcolsep}{2.5pt}

\begin{tabular}{lccc|cc}
\toprule 
& \makecell{Cross-Tile\\Attention}
& \makecell{Context-Aware\\Decoding}
& \makecell{Global\\Alignment}
& LPIPS $\downarrow$
& SSIM $\uparrow$ \\
\midrule

(a) 
& & & &
0.216 & 0.736 \\

(b)
& \checkmark & & &
0.211 & 0.737 \\

(c)
& \checkmark & \checkmark & &
0.212 & 0.747 \\

(d)
& \checkmark & \checkmark & \checkmark &
\cellcolor{best}\textbf{0.206} &
\cellcolor{best}\textbf{0.749} \\

\bottomrule
\end{tabular}
\end{table}

Table~\ref{tab:ablation} quantitatively evaluates the contribution of each component. Adding cross-tile attention improves LPIPS from 0.216 to 0.211
and SSIM from 0.736 to 0.737. Incorporating context-aware decoding further
increases SSIM to 0.747 while maintaining a comparable LPIPS of 0.212,
indicating improved structural similarity without degrading overall
perceptual quality. Finally, global appearance alignment achieves the best
LPIPS of 0.206 while maintaining the highest SSIM of 0.749. These results
show that the proposed components provide complementary improvements to
the quality of the generated scenes.

Figure~\ref{fig:ablation} presents a qualitative ablation study on the Town scene from the synthetic subset of UrbanScene3D. Compared with the baseline, incorporating cross-tile attention reduces visible seams by allowing each tile to interact with its neighboring tiles during generation. Adding context-aware decoding further suppresses boundary artifacts by incorporating neighboring context during tile-wise decoding. Finally, global alignment improves consistency beyond adjacent regions: spatially distant tiles exhibit more similar ground patterns and color characteristics, demonstrating improved global appearance consistency across the scene.

\section{Conclusion} 

We presented a training-free framework for large-scale textured mesh generation from multi-view images using pretrained 3D generative priors. Our framework extends tiled generation with adaptive scene decomposition and local and global coordination across spatial regions. Specifically, it incorporates cross-tile context during generation and decoding to preserve geometric continuity and aligns texture features across distant regions to improve global appearance consistency. Experiments on diverse large-scale scenes demonstrate improved geometric and appearance fidelity over existing approaches while enabling fine-grained generation of large-scale textured meshes. Our work provides a practical approach to scaling pretrained 3D generative priors beyond their original object-centric domain without additional training.

\bibliographystyle{plain}
\bibliography{main}

\end{document}